\documentclass{article}

\usepackage[preprint]{neurips_2023}

\usepackage[utf8]{inputenc} 
\usepackage[T1]{fontenc}    
\usepackage{hyperref}       
\usepackage{url}            
\usepackage{booktabs}       
\usepackage{amsfonts}       
\usepackage{nicefrac}       
\usepackage{microtype}      
\usepackage{xcolor}        
\usepackage{graphicx}
\usepackage{enumitem}
\usepackage{natbib}
\usepackage{tabularx}
\title{Harnessing Business and Media Insights\\with Large Language Models}

\author{
       Yujia Bao$^{1*}$
  \And Ankit Parag Shah$^{1*}$
  \And Neeru Narang$^{2*}$
  \AND Jonathan Rivers$^3$
  \And Rajeev Maksey$^3$
  \And Lan Guan$^1$
  \And Louise N. Barrere$^1$
  \And Shelley Evenson$^2$
  \And Rahul Basole$^2$
  \And Connie Miao$^1$
  \And Ankit Mehta$^1$
  \AND Fabien Boulay$^1$
  \And Su Min Park$^1$
  \And Natalie E. Pearson$^2$
  \And Eldhose Joy$^2$
  \AND Tiger He$^1$
  \And Sumiran Thakur$^2$
  \And Koustav Ghosal$^2$
  \And Josh On$^2$
  \AND Phoebe Morrison$^2$
  \And Tim Major$^2$
  \And Eva Siqi Wang$^2$
  \And Gina Escobar$^2$
  \AND Jiaheng Wei$^1$
  \And Tharindu Cyril Weerasooriya$^1$
  \And Queena Song$^2$
  \And Daria Lashkevich$^2$
  \And Clare Chen$^1$
  \And Gyuhak Kim$^1$
  \And Dengpan Yin$^1$
  \And Don Hejna$^2$
  \And Mo Nomeli$^2$
  \AND Wei Wei$^1$
  \INST $^1$Center for Advanced AI, Accenture
  \Inst $^2$Accenture
  \Inst $^3$Fortune Media
  }
\begin{document}

\maketitle

\begin{abstract}
This paper introduces Fortune Analytics Language Model (\texttt{FALM}).
\texttt{FALM} empowers users with direct access to comprehensive business analysis, including market trends, company performance metrics, and expert insights. Unlike generic LLMs, \texttt{FALM} leverages a curated knowledge base built from professional journalism, enabling it to deliver precise and in-depth answers to intricate business questions. Users can further leverage natural language queries to directly visualize financial data, generating insightful charts and graphs to understand trends across diverse business sectors clearly. \texttt{FALM} fosters user trust and ensures output accuracy through three novel methods: 1) Time-aware reasoning guarantees accurate event registration and prioritizes recent updates. 2) Thematic trend analysis explicitly examines topic evolution over time, providing insights into emerging business landscapes. 3) Content referencing and task decomposition enhance answer fidelity and data visualization accuracy.
We conduct both automated and human evaluations, demonstrating \texttt{FALM}'s significant performance improvements over baseline methods while prioritizing responsible AI practices. These benchmarks establish \texttt{FALM} as a cutting-edge LLM in the business and media domains, with exceptional accuracy and trustworthiness.
\end{abstract}
\clearpage

\section{Introduction}
The ever-shifting business landscape necessitates powerful tools for staying ahead. While large language models (LLMs) have shown promise in analyzing vast datasets and generating valuable insights for market analysts~\citep{wu2023bloomberggpt}, they haven't yet reached the full potential of state-of-the-art general-purpose LLMs~\citep{achiam2023gpt,team2023gemini,jiang2024mixtral,touvron2023llama}. This gap stems from the unique challenge of integrating diverse data sources and modalities within the business domain, while ensuring exceptional accuracy and fidelity for real-world applications.

This paper introduces Fortune Analytics Language Model (\texttt{FALM}), a business-centric AI system designed specifically for the business and media domain. We leverage the rich heritage of Fortune Media, a globally renowned publication known for its in-depth feature articles spanning over 95 years. This vast knowledge base, created by professional journalists, forms the foundation for \texttt{FALM}'s training data.  By focusing on the business and media domain, \texttt{FALM} empowers users with intuitive and insightful business analysis through two core functionalities:

\begin{itemize}[leftmargin=20pt]
    \item \textbf{Business-Centric Question Answering:} \texttt{FALM} leverages diverse data sources, including news articles, video interviews, ranking lists, financial metrics, and business leader profiles, to answer complex questions about the ever-evolving business landscape. It identifies trends across various topics, from market fluctuations to industry leadership shifts, and offers its analysis based on financial indicators. For instance, if a user inquires, ``\emph{How has the interest in sustainable consumption among younger consumers affected purchasing patterns?}'' \texttt{FALM} will utilize recent news articles, reports, and video interviews to generate a thorough and precise analysis.
    \item \textbf{Data Visualization:}     \texttt{FALM}'s capabilities extend beyond question answering. Users can directly visualize financial data through natural language queries and ask for charts or graphs to gain a more intuitive understanding of complex financial trends. For instance, a user may ask: ``\emph{Can you show me a graph comparing the revenue of the top 10 companies over the last decade?}'' \texttt{FALM} would interpret the request and generate an easy-to-understand chart illustrating the revenue of the top performers over the past ten years.
\end{itemize}

However, developing such a tool presents its challenges. While LLMs have revolutionized many domains~\citep{saab2024capabilities,george2023review,webersinke2021climatebert,chen2024octopus}, ensuring the correctness and consistency of LLM-generated outputs, especially for complex business reasoning tasks, remains an active area of research. This paper addresses three key challenges in developing business-centric AI systems, focusing on real-world user requirements.

\paragraph{Time-Aware Reasoning} 
Value of business data is often contingent on its timeliness, and accurate contextualization of events necessitates a robust anchoring to relevant temporal coordinates. For instance, if a 2020 news article references an event from "last year," the system should interpret "last year" as 2019, not 2023. Furthermore, when users pose queries on a certain topic, the model should prioritize presenting the most current news and information. \texttt{FALM} integrates an understanding of time into the decision-making processes through instruction finetuning. As a result, the model is able to maintain relevance, reliability, and accuracy in its responses, enhancing user satisfaction and trust.

\paragraph{Thematic Modeling for Trend Analysis}
Business-related queries encompass a diverse range of current affairs, corporate activities, social issues, as well as market and industry trends reported by journalists. It is important for the model to comprehend the chronological progression of various topics or "themes" over time in order to appropriately answer questions such as ``\emph{How has AI evolved in the last five years?}'' \texttt{FALM} explicitly models trends across several temporal scales (month, quarter, year, multi-year) and provides a comprehensive view of how a particular topic has evolved over time, taking into account short-term fluctuations as well as long-term trends. 

\paragraph{Accuracy and Trust} Accuracy is critical when it comes to business and financial data. As probabilistic models, LLMs introduce inherent scholastic behaviors, making precise control of accuracy challenging. We address accuracy through three main approaches:
\begin{itemize}[leftmargin=20pt]
    \item Task Decomposition: We decompose many of the challenging tasks into stages in order to achieve control over accuracy. For example, in the financial data visualization task, we decompose chart generation into code generation, followed by an execution phrase using a financial spreadsheet. This ensures data fidelity, as the generated code leverages verified financial metrics as to produce the visualizations.

    \item Knowledge Boundaries: Unlike general-purpose LLMs, our model responds to queries within a predefined business domain.. Suppose a question falls outside the training data's coverage; in that case, our model generates a "rejection response" that declares that the inquired topic lies outside of its designed domain instead of potentially generating ungrounded outputs.

    \item Content Referencing: We integrated a content referencing system into the interface to provide users with insights into the model's reasoning process. This feature offers users insight into the model's decision-making process by providing hyperlinks within the generated text or in summarized articles displayed alongside the responses.

\end{itemize}

Safety is of paramount importance to AI systems. We employ a multi-layered approach to ensure the model generates safe and reliable outputs:

\begin{itemize}[leftmargin=20pt]
    \item Corpus guardrail: To ensure the reliability of the information that \texttt{FALM} generates, we implement a multi-layered corpus guardrail system during the data preparation process to mitigate bias and protect privacy.
    \item Safety alignment: This process involves training the model on a curated dataset with unsafe or misleading prompts. By exposing \texttt{FALM} to these questions, the model learns to identify and reject these harmful queries at inference time.
    \item Deployment guardrail: Even after training, it is crucial to maintain safeguards against potential risks during deployment. We implement robust deployment guardrails to ensure the system operates safely and protects user privacy. 
    \item Continuous Human Feedback: During the development process, we have continuously collected user feedback via surveys and testing sessions. This feedback process allows us to identify and rectify potential safety concerns early on, enabling us to constantly refine the model and ensure it operates within safe boundaries.
\end{itemize}

In the following sections, we will describe our training data, business-centric question-answering capabilities, data visualization techniques, the content reference system, safety guardrails, and system designs. We will also present evaluation benchmarks to assess the performance of each \texttt{FALM} components.

\section{Fortune Knowledge}

\subsection{Articles}

\texttt{FALM} is trained on written articles published by Fortune Media, both online and in print. The training data covers a wide array of content, including short-form news articles, long-form investigative pieces, opinion editorials, and in-depth analyses, equipping the model with a deep understanding of the business world, financial markets, and the global economy.

The article corpus spans a diverse set of themes drawn from Fortune's carefully selected and organized sections including the following.

\begin{itemize}[leftmargin=20pt]
    \item \textbf{Finance:} Content focusing on financial markets, investing, banking, personal finance, wealth management, and global economic trends.
    \item \textbf{Leadership:} Articles exploring various aspects of leadership, management strategies, and the experiences of successful business leaders.
    \item \textbf{Success:} Stories highlighting the achievements of individuals, companies, and organizations across various industries.
    \item \textbf{Tech:} Pieces related to the latest advancements in artificial intelligence, cybersecurity, blockchain, startups, and the tech industry as a whole.
    \item \textbf{Asia and Europe:} Content focused on business, economics, and technology in these specific regions.
    \item \textbf{Environment:} Articles discussing climate change, renewable energy, conservation efforts, and the intersection of business and the environment.
    \item \textbf{Fortune Crypto:} Insights into the world of cryptocurrencies, blockchain technology, and their impact on finance and business.
    \item \textbf{Health and Wellness:} Content related to healthcare technology, the pharmaceutical industry, healthcare policy, and trends in medicine, as well as personal health and wellness.
    \item \textbf{Retail and Lifestyle:} Articles exploring the retail industry, consumer trends, travel, real estate, luxury goods, and entertainment.
    \item \textbf{Politics:} Pieces analyzing the intersection of politics, business, and economics, as well as their impact on various industries.
\end{itemize}

The Fortune articles corpus used in our model extends beyond traditional news articles to encompass a diverse range of Fortune content. This includes categories like Newsletters, Features, and Commentary, offering in-depth analysis and expert perspectives. Additionally, the corpus incorporates coverage of major events like the Most Powerful Women (MPW) conference and the CEO Initiative, providing valuable insights into leadership trends.  It also integrates Personal Finance content and Fortune's recommended resources, equipping users with practical financial knowledge. This comprehensive collection fosters a well-rounded understanding of the multifaceted world of business and finance.

\subsection{Video Transcripts}
In addition to written articles, Fortune provides video content through its premium feature, Fortune On-Demand. This service offers exclusive access to:
\begin{itemize}[leftmargin=20pt]
    \item \textbf{News:}  Timely and relevant business news videos covering the latest developments and trends.
    \item \textbf{Interviews:} Conversations with CEOs, entrepreneurs, and other influential figures in the business world.
    \item \textbf{Series:}  Engaging video series that delve into specific topics or industries, offering comprehensive insights.
    \item \textbf{Conferences:} Live streams, highlights, and expert discussions from prestigious conferences and events.
    \item \textbf{Insights:} Data-driven analysis and expert opinions on key business issues, delivered through informative videos.
\end{itemize}

Video transcripts offer an engaging and interactive experience for users. However, the conversational nature of transcripts makes them less structural compared to articles, with the possibility of disfluencies and incomplete sentences arising from their video origins. To mitigate this issue, we pass video transcripts into LLMs and summarize them before they are consumed in the model training process.  Building upon research in transcript summarization~\citep{murray2005extractive,li2017adversarial,li2017multi,fajtl2019summarizing,radford2023robust,krubinski2023mlask,lin2023mm}, our approach refines LLM-generated summaries with human verification. This process transforms the unstructured conversational video content into a more organized and consumable format for model training purposes.

\subsection{Company Ranking Lists}
Fortune magazine publishes several well-known ranking lists, each focusing on different aspects of the business world. Two prominent instances are:
\begin{itemize}[leftmargin=20pt]
    \item \textbf{Global 500:} The Fortune Global 500 list is an annual ranking of the world's biggest businesses, providing a snapshot of their performance as of a specific date. Companies on the most recent list have reported financial data for their fiscal year ending on or before March 31st of the previous year. To be eligible, companies must be publicly traded and release their financial information. The Global 500 is a valuable indicator of which industries are thriving globally.
    \item \textbf{Fortune 1000:} The Fortune 1000 is an annual list published by Fortune magazine that ranks the 1,000 largest companies in the United States based on their total revenue. The list offers valuable insights into the health and trends of the U.S. business landscape, showcasing the most influential companies and serving as a useful tool for understanding the U.S. economy.
\end{itemize}

In addition to the pure rankings, both lists contain detailed financial metrics to provide additional insights regarding the listed companies including:
\begin{itemize}[leftmargin=20pt]
    \item \textbf{Company Details:} Each company's official name, rank based on total revenue, the year it was founded, and associated industry, sector, and more. Industry information provides insight into trends, shifts, and the emergence of new sectors, such as the rise of technology and digital services or the changes of the manufacturing sector.
    \item \textbf{Financials:} This data captures key financial metrics including revenue, profits (net income after taxes and expenses have been deducted), assets, and market value. This information is critical to the understanding of the economic impacts companies bring globally.
    \item \textbf{Geographic Details:} Geographic details contain the country and region where companies are headquartered, which is important for understanding regions that are economically strong and growing for investment purposes. The presence of a company in multiple countries also provides information about its expansion beyond domestic markets.
    \item \textbf{Employment Information:} This data documents the number of employees for each company, offering insight into large employers and distribution across geographical regions.
    \item \textbf{Performance Metrics:} Year-over-year change in revenue provides insights into which companies and sectors are expanding or contracting. Earnings per share (EPS) also offers information about the financial health and operational efficiency of the companies.
\end{itemize}
In addition to these two rankings, Fortune regularly publishes lists in various realms such as human resources, technology, and leadership. Examples of these lists include: ``World's Most Admired Companies,'' ``40 Under 40,'' ``Fortune Most Powerful Women Entrepreneurs,'' ``100 Best Companies to Work For,'' ``100 Fastest Growing Companies,'' ``The Unicorn List,'' ``Businessperson of the Year,'' ``Change the World,'' ``The World's 50 Greatest Leaders,'' and ``Future 50.''

Overall, Fortune has published 263 such lists over the past few decades. These rankings provide a great anchor point for a business-centric AI to analyze market trends and in order to generate insights and analyses. The vast amount of data in these lists can help businesses make informed decisions and stay competitive in their respective industries.

\section{Business-centric Question Answering}\label{sec:business-centric-qa}

\texttt{FALM} tackles knowledge comprehension across various information sources from the business and media domain.  We can categorize \texttt{FALM}'s QA functionality into five essential subtasks, each focusing on a specific information need.
\begin{itemize}[leftmargin=20pt]
    \item \textbf{Article QA:} The objective of Article QA is to train the model in understanding and providing detailed answers to user questions related to specific articles within the Fortune data collection. Example questions include, ``\emph{Can you provide more information about Fortune's partnership with Accenture on their latest AI tool?}''
    \item \textbf{Topic QA:} In many instances, users are less concerned with the specifics of individual events and more interested in overall developments within a particular field or timeframe. In Topic QA, we prioritize uncovering trends and thematic progressions within the business domain. An example question would be, ``\emph{What was Fortune's view on inflation in April 2024?}'' \texttt{FALM} would leverage its thematic modeling capabilities to analyze relevant articles and identify key themes and their evolution over time.
    \item \textbf{Metric QA:} This subtask focuses on answering questions related to quantifiable business metrics. For instance, users may want to understand key metrics such as revenue for each company by asking questions like, ``\emph{What was Walmart's revenue in 2024?}'' Here, carefully designed knowledge boundaries are crucial to ensure the model stays within its training domain instead of hallucinating. This is achieved by systematically developing "rejection QA pairs" - essentially training examples where the model learns to identify questions it cannot answer accurately. Imagine a user asks, ``\emph{What was the average stock price of Apple in 2025?}'' \texttt{FALM} would reject this user question.
    \item \textbf{Ranking QA} Fortune regularly publishes ranked lists to offer subscribers industry insights. Examples include ``World's Most Admired Companies,'' ``100 Fastest Growing Companies,'' and ``40 under 40.'' Ranking List QA aims to integrate these rankings seamlessly into the \texttt{FALM} model. Similar to Metric QA, \texttt{FALM} would leverage its knowledge boundaries to ensure the retrieved ranking information is relevant and reliable.
    \item \textbf{Persona QA} Our aim is to develop a distinct model persona for \texttt{FALM}, enabling subscribers to comprehend the underlying design logic through simple questions like, ``\emph{Are there any philosophical principles embedded in your programming?}''
\end{itemize}

To ensure users receive the most relevant and engaging information, FALM integrates time-aware reasoning (discussed earlier) into all functionalities. This ensures the model prioritizes the latest updates and considers the temporal context of retrieved information.

We understand that generic answers, while informative, can lack the depth and engagement users desire. Research suggests that detailed and context-rich responses are crucial for user satisfaction~\citep{zhao2020knowledge,dinan2018wizard}. To address this, \texttt{FALM} is designed specifically to provide detailed responses that incorporate relevant examples whenever possible. It leverages the vast archive of Fortune's business news coverage to draw upon specific and illustrative examples that support its claims.  By offering actionable insights and relevant information embedded within concrete examples, \texttt{FALM} aims to significantly improve user engagement and satisfaction.

Despite these distinct subtasks and different knowledge sources, \texttt{FALM} employs a unified instruction fine-tuning framework for overall efficiency. Building high-quality instruction finetuning datasets is challenging. Existing approaches for developing high-quality in-domain question-answering datasets often require extensive human annotation~\citep{rajpurkar2016squad,trischler2016newsqa,MultiRC2018,rajpurkar2018know, jin2019pubmedqa, pal2022medmcqa}. However, this approach is not feasible due to the following reasons.

\begin{itemize}[leftmargin=20pt]
    \item Given the extensive timespan of Fortune knowledge, which dates back to the last century, employing human annotation would be an incredibly labor-intensive endeavor.
    \item While question generation can be streamlined with appropriate guidance for annotators, answer generation presents a greater challenge, which requires meticulous synthesis of information, adherence to Fortune's journalistic style, and rigorous quality control to ensure accuracy and consistency.
\end{itemize}

We tackle this challenge and break down each subtask using LLM agents~\citep{wang2024survey,cai2023large,du2023improving,hong2023metagpt,zhuge2023mindstorms,park2023generative,akata2023playing}. Specifically, we designate specific LLM agents for each task and input raw Fortune knowledge into our data generation pipeline, effectively automating the time-consuming manual annotation process. We also incorporate human intervention with the LLMs to guarantee that the generated data accurately reflects the original Fortune knowledge and aligns with our stylistic expectations.

\subsection{Evaluation}
In this section, we conduct a comprehensive analysis of the model's performance on business-centric question answering. We employ two evaluation methodologies: comparative evaluation and independent evaluation.

\paragraph{Comparative Evaluation} This evaluation involves comparing responses generated by different models and evaluating their corresponding win rates. Comparative evaluation is beneficial for tasks where there is no golden ground truth, such as answering the question ``\emph{Can you tell me more about recent news on inflation?}'' Human judgment plays a crucial role in this process as human evaluators compare answers from various models based on fluency, accuracy, and relevance \citep{ribeiro2020beyond}. To complement human evaluation, we use LLM-based evaluation tools such as Prometheus \citep{kim2024prometheus}, which leverage language models to assess text quality. This approach allows for systematic evaluations at scale \citep{celikyilmaz2020evaluation,hackl2023gpt}. However, a drawback is that LLM-based tools may favor model responses similar to their training corpus or style \citep{dubois2024length}. Our evaluation pipeline integrates both LLM-based and human evaluators for a balanced approach between scalability and human alignment~\citep{gehrmann2021gem}.

\paragraph{Independent Evaluation} In this method, we analyze the response from one model and evaluate its performance based on predefined metrics or ground truths. For instance, in Metric QA, we will assess the exact match of requested metrics. In data visualization, we will determine if the output chart aligns with user queries. Independent evaluation can also be categorized into three primary categories: statistical measures, human-defined rubrics, and LLM-based or human evaluations. Statistical measures such as BLEU \citep{papineni2002bleu}, ROUGE \citep{lin2004rouge}, and METEOR \citep{banerjee2005meteor} are commonly used to assess the similarity between generated text and reference text. Similarly, LLM-based evaluation and human evaluators with customized rubrics (i.e., fluency, coherence, relevance, and factual accuracy) \citep{ribeiro2020beyond,lin-etal-2022-truthfulqa} are also widely adopted in independent evaluations. Human judgment remains crucial for assessing text quality based on factors such as fluency, accuracy, relevance, and truthfulness.

\begin{figure}
    \centering
    \includegraphics[width=0.9\linewidth]{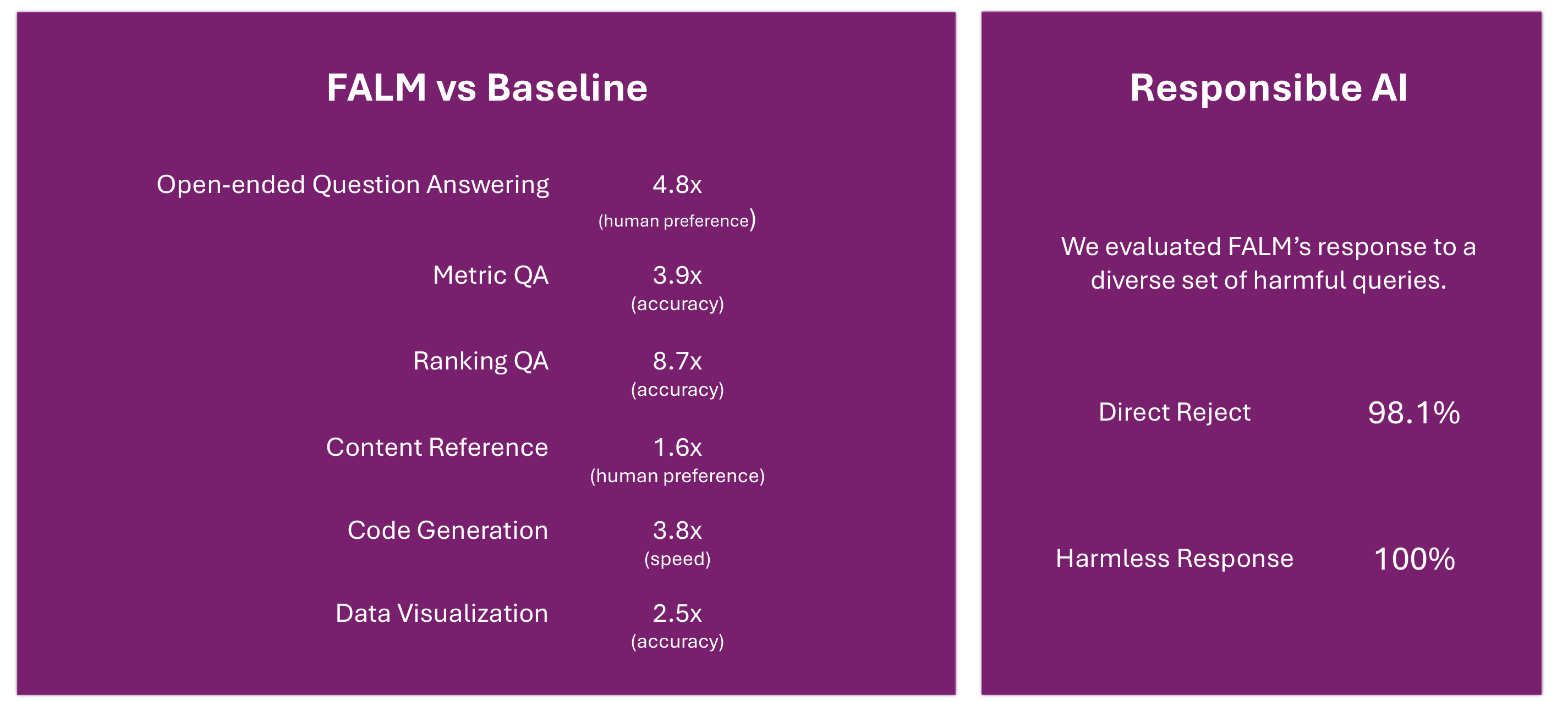}
    \caption{
    This figure summarizes \texttt{FALM}'s performance across various tasks compared to a state-of-the-art open-source LLM with prompt engineering as the baseline. \texttt{FALM} achieves significant improvements in: Open-Ended Question Answering (4.8x), Metric QA (3.9x), Ranking QA (8.7x), Content Referencing (1.6x), Data Visualization (2.5x). Notably, \texttt{FALM}'s focus on the business and media domain also leads to a substantial increase in inference speed (3.8x).  Furthermore, \texttt{FALM} demonstrates strong performance in handling harmful prompts, ensuring responsible AI practices.}\label{fig:patha_eval}
\end{figure}

\paragraph{Open-ended questions}
We assess the performance of our model in addressing open-ended business-related queries by utilizing a dataset of prompts gathered from real-world users. These prompts are categorized into six categories:
\begin{itemize}[leftmargin=20pt]
\item \textbf{American (12\%)}: This category comprises questions about U.S.-based companies. Example prompts include: ``\emph{What's fueling Valero Energy's growth?}'' and ``\emph{What emerging trends exist in large-cap U.S. Companies?}''
\item \textbf{Global (12\%)}: The questions in this category focus on businesses outside the United States. An example is: ``\emph{What is Saudi Aramco's business model?}``
\item \textbf{Competition (17\%)}: This category covers questions about competition and relationships among different companies. An illustrative question would be: ``\emph{How is American Express distinguishing itself from its competitors?}''
\item \textbf{Leader (16\%)}: The questions in this category asks about business leaders. A sample prompt reads: ``\emph{What was Mark Cuban's career trajectory?}''
\item \textbf{News and topics (43\%)}: This category encompasses queries about recent news topics. Example questions are: ``\emph{How has the landscape for wind energy changed recently?}'', ``\emph{How have different generations' attitudes towards savings and investing evolved over time?}'', ``\emph{How can we redefine work in the age of Gen AI?}''
\end{itemize}

Our model is compared against state-of-the-art open-source LLMs with prompt engineering. During the evaluation, human raters remain oblivious to the identity of the backend LLMs that generate the responses. The raters are tasked with ranking both responses based on their accuracy, redundancy, level of detail, relevance, tone, and style. Each question receives input from three annotators, and the majority vote determines the final label. The results, displayed in Figure~\ref{fig:patha_eval}, indicate that \texttt{FALM} outperforms the baselines by 4.8 times.

\paragraph{Metric QA} Our benchmark includes a diverse set of unseen questions about the financial metrics of top companies from Global 500 and Fortune 1000. The breakdown is as follows: 27\% on revenue and 25\% on profits, with the rest covering assets, earnings, market value, and number of employees. We evaluate model responses against ground truth metrics. Our baseline for comparison is a model variant trained without the Metric QA dataset (though it still accesses metric information during continuous pre-training). The following rubrics define accuracy:
\begin{itemize}[leftmargin=20pt]
    \item  If users request a company's metric at a specific year within the knowledge cutoff, answers must match ground truth exactly.
    \item For requests outside the cutoff, the answer should clarify unavailability and provide the latest data as reference.
\end{itemize}
Comparing to the baseline, \texttt{FALM} improves the Metric QA performance by 3.9 times.

\paragraph{Ranking QA}
We developed a benchmark consisting of a diverse set of unseen questions, focusing on ranking list information. This diverse set covers various language types and styles. Model responses are evaluated against ground truth lists. To demonstrate our model's effectiveness, we conducted an ablation study comparing it to a baseline without data generated by the Ranking QA dataset pipeline during instruction fine-tuning (though the baseline still has access to ranking information in continuous pre-training).

We use the following rubrics to determine accuracy:
\begin{itemize}[leftmargin=20pt]
    \item If users request the top k of a ranking list for a specific year, the answer must align with the ground truth.
    \item For questions asking only for the top companies or people on a ranking list for a given year, we consider answers correct if they return the correct top 5 or top 10.
    \item If users request a ranking list for a year when that information is unavailable, we consider an answer correct if it redirects users to the closest available ranking list.
\end{itemize}
Comparing to the baseline, \texttt{FALM} improves the Ranking QA performance by 8.7 times.

\section{Data Visualization}
Visualizing key financial metrics across top companies over decades is crucial for a business-centric AI. This capability offers users valuable insights into the economic trends of various industries or countries. We consider three primary types of plots, each with a distinct focus on the insights they convey.

\begin{itemize}[leftmargin=20pt]
    \item \textbf{Bar plot:} Bar plots facilitate easy comparison between different categories or groups. A simple example might be: ``\emph{Plot the revenue for Apple, Google and Nvidia in 2024.}'' This visualization provides a clear representation of discrete data and is ideal for displaying categorical data or data with distinct categories.
   \item \textbf{Line plot:} Line plots excel at showing trends or patterns over time. For instance, users can request: ``\emph{Show me the revenue for Apple, Google and Nvidia since 2014.}'' This type of plot effectively highlights changes and variations in data over time or other continuous variables.
  \item \textbf{Scatter plot:} Scatter plots are effective for visualizing relationships between two variables. This is useful for assessing the strength and direction of the relationship between variables. For example, to study how revenue relates to the number of employees, users can ask: ``\emph{Compare the revenue and the number of employees for the top 10 companies on the Fortune 1000 list.}''
\end{itemize}

As demonstrated in these examples, queries often require a significant amount of reasoning over raw company financial metrics, such as filtering, aggregation, or ranking. While LLMs have shown remarkable performance in similar tasks, like chain-of-thought reasoning \citep{wei2022chain,kojima2022large,wang2023selfconsistency}, rationale engineering \citep{fu2022complexity,zhang2023automatic,li2022advance}, bootstrapping and self-improving \citep{zelikman2022star,huang2023large}, combining the fact that there are also many numerical metrics can make directly asking the LLM to generate visualization challenging.

To ensure visualization accuracy and data fidelity, we break down the task into two steps. First, \texttt{FALM} generates Python code that retrieves company metrics from an external dataframe. This code then performs the necessary reasoning and plotting using standard data frame manipulation techniques. This decomposition offers several advantages:

\begin{itemize}[leftmargin=20pt]
    \item \textbf{Data Grounding:}  By relying on the external dataframe for data, the generated visualizations are always based on the correct company information. There is no need for the LLM to memorize specific values.
    \item \textbf{Time-Aware Reasoning:} The model can adapt to user requests regarding timeframes. It can retrieve financial data for specific years (e.g., ``\emph{revenue of Alphabet for 2012, 2015, and 2020}'') or prioritize recent data when no year is specified (e.g., ``\emph{plot the revenue of Apple}'').
    \item \textbf{Scalability:} Updates to company financials only require modifying the external dataframe, not retraining the model. The generated code remains functional.
    \item \textbf{Customization flexibility:} Since the model outputs Python code, visualizations can be customized during execution to incorporate user-friendly features independent of the model itself.
\end{itemize}

While LLMs are revolutionizing software development by assisting with code completion and translation of natural language descriptions into functional code~\citep{roziere2023code,luo2023wizardcoder,li2023starcoder,lozhkov2024starcoder,achiam2023gpt,reid2024gemini}, ensuring the correctness and user-alignment of LLM-generated code, especially for complex tasks like financial data visualization, remains a challenge~\citep{chen2021evaluating,ziegler2019fine}.

We propose a code generation pipeline that leverages LLMs agents for creating the instruction finetuning data. Here, we address correctness and robustness concerns through a two-step process:

\begin{itemize}[leftmargin=20pt]
    \item \textbf{Synthesized Code Generation:}  We develop a template-based system to ensure the generated visualization code is structurally sound. This system offers diversity by covering a broad range of data manipulation procedures suitable for financial data visualization tasks.
    \item \textbf{LLM-driven Natural Language Conversion:}   LLMs are then utilized to translate these pre-defined code templates into natural language questions. We can then utilize these question-code pairs for instruction finetuning.
\end{itemize}

By combining these steps, we aim to achieve a balance between code correctness and the flexibility offered by LLM-based generation.

\subsection{Evaluation}
Evaluating code generation involves accessing multiple dimensions, including correctness, quality, and efficiency. Existing benchmarks provide a structured approach to this evaluation. For instance, the CodeXGLUE benchmark encompasses a variety of tasks such as code summarization, completion, and translation, using metrics like BLEU, ROUGE, and accuracy to evaluate the syntactic and semantic quality of the generated code \citep{lu2021codexglue}. Correctness is rigorously tested using datasets like HumanEval, where generated code undergoes unit tests that verify its functional accuracy by ensuring it produces the expected outputs for given inputs \citep{chen2021evaluating}. It enables a detailed assessment of whether the code meets the specified problem requirements. Furthermore, CodeNet \citep{puri2021codenet} offers an extensive collection of coding problems and solutions across multiple programming languages, facilitating the evaluation of not only correctness but also the model's versatility and robustness in handling diverse coding tasks. Finally, evaluating downstream performance, i.e., the practical execution of generated code, is another popular evaluation direction among more recent works \citep{zhou2023codebertscore,ni2023l2ceval,dong2023codescore}.

We benchmark the performance of our LLM-based code generation and code execution framework for data visualization tasks. We conduct independent evaluation and benchmark the generated chart from the model based on ground truth. We consider the following evaluation metrics:
\begin{itemize}[leftmargin=20pt]
    \item \textbf{Code Execution Rate:} This metric measures the percentage of generated code that executes successfully without errors. A high execution rate indicates the model's ability to produce functional code.
    \item \textbf{Data Match Accuracy:} We execute the generated code and compare its output with the ground truth results of the data manipulations it performs. This ensures the code accurately transforms the data as intended.
\end{itemize}

\paragraph{Benchmark dataset}
To evaluate our model's performance comprehensively, we created our benchmark leveraging two sources: a templated prompts dataset and a free-form prompts dataset.

The templated prompts dataset consists of prompts derived from a set of pre-defined code templates. These templates encompass a variety of chart types and filter options. Notably, the template design is independent of our code generation pipeline to prevent information leakage.

Real-world user prompts often deviate from pre-defined templates. To understand the model's behavior under these conditions, we introduce a dataset of 7,860 free-form prompts. This dataset captures the natural distribution of user requests and was collected through a user survey. During the survey, participants were presented with two options:
\begin{itemize}[leftmargin=20pt]
    \item Free-Form Question: Users were asked to provide a free-form question about something they were interested in visualizing.
    \item Instruction for Existing Chart: Users were shown a pre-defined chart and asked to write instructions for the AI to generate that specific chart.
\end{itemize}
This approach resulted in a dataset containing 7,860 user queries (3,727 free-form text prompts and 4,133 instruction prompts).

\paragraph{Code Execution Rate:}
This step evaluates the ability of the generated code to execute successfully and produce the desired chart. We analyze errors encountered during execution and categorize them into two primary types:
\begin{itemize}[leftmargin=20pt]
\item Syntax Errors: These errors indicate incomplete or syntactically incorrect code, such as unmatched parentheses, missing colons, or improper indentation.
\item Runtime Errors: These errors occur during code execution despite successful syntax checks. Examples include division by zero, missing column names, or duplicate index errors.
\end{itemize}
We benchmark the performance of \texttt{FALM} against a state-of-the-art LLM with prompt-engineering. Despite significant prompt engineering effort, the baseline achieves a success rate of only 75.25\% across all four chart types. At the chart-type level, the baseline model achieves an accuracy of 87\% for bar plots, 64\% for line plots, and 70\% for scatter plots. In contrast, our model demonstrates significantly higher robustness, achieving a success rate of 99.3\%. At the chart-type level, our model achieves an accuracy of 100\% for bar and scatter plots and 98.10\% for line plots.

Our model also achieves a significant 3.8x speedup in inference latency compared to the base model with prompt engineering. This is due to its optimized parameter efficiency and the elimination of lengthy prompt appending at runtime. This translates to substantial improvements in throughput, productivity, and cost savings, particularly for large-scale deployments. Faster inferences deliver quicker results to end-users, enhancing user experience and satisfaction. This is especially crucial for real-time applications where latency directly impacts performance.

\paragraph{Data Match Accuracy:}
This evaluation step assesses the quality of the data plotted in the visualizations generated by LLM code. It focuses on two key aspects:
\begin{itemize}[leftmargin=20pt]
    \item Data Consistency:  Verifies that the data points plotted in the chart accurately reflect the user's intended output. In other words, it ensures the LLM code retrieves and visualizes the correct data.
    \item Data Validity:  Ensures the generated plot does not contain any non-existent data points.
\end{itemize}
To achieve this, we compared the results produced by \texttt{FALM} with the ground truth, representing the ideal data output based on user query. The comparison focused on two key areas:
\begin{itemize}[leftmargin=20pt]
\item Chart Axes: This compares the columns chosen for the x-axis and y-axis in both plots.
\item Data Values: This compares the actual data points plotted on the axes, including the total number of data rows.
\end{itemize}

Comparing to the prompt engineering baseline, \texttt{FALM} improves the data manipulation accuracy by 2.5 times.

\section{Content Reference}
LLMs are revolutionizing information access and content creation. However, the opaque nature of their knowledge provenance poses a challenge to responsible use, making the integration of citation practices within LLM outputs essential. Citations offer both technical advantages and ethical considerations for LLMs:

\begin{itemize}[leftmargin=20pt]
    \item \textbf{Improved Interpretability:} Citations provide context for the LLM's outputs, enabling users to verify the information and understand the rationale behind the LLM's reasoning by examining the cited sources. This fosters interpretability and builds trust in the LLM's capabilities.
    \item \textbf{Attribution and Credit:} \texttt{FALM} is built on top of Fortune's knowledge accumulated over the past few decades. Citations ensure proper attribution and credit to the original editorial team, upholding ethical principles of intellectual property.
    \item \textbf{Mitigating Disinformation:} The spread of misinformation is a growing concern. Citations empower users to evaluate the credibility of information presented by the LLM by referencing the original sources. This promotes responsible information consumption and helps mitigate the risks associated with disinformation.
\end{itemize}

Directly generating citations during LLM generation is challenging, as the model must remember all links to the original knowledge. Additionally, this method is unreliable due to the randomness involved in potential hallucinations during LLM generation. Following \cite{huo2023retrieving}, we generate content references by directly searching the answer produced by the model across the entire Fortune knowledge base. This ensures that the returned links are always consistent with the model's answer \citep{nogueira2019passage,guu2020retrieval,lin2022pretrained,zhang2023extractive}.

Our content reference pipeline comprises two components: retrieval and re-ranking. The retrieval stage aims to eliminate irrelevant information from the Fortune knowledge base, retaining only the top articles. Subsequently, in the re-ranking step, we scrutinize the retrieved information in greater detail to identify the best match(es).

\paragraph{Evaluation}
To evaluate the effectiveness of our content reference system in retrieving relevant information for user queries, we employed human annotations. Annotators were presented with a question, the system's generated answer, and a corresponding article retrieved through the re-ranking process. Using a binary scale, they assessed whether the article provided substantial supporting evidence for the answer. To mitigate potential bias and ensure accuracy, each question-answer-article set was evaluated by three independent annotators. The final label was determined by majority vote.

We compared the performance of our system with a state-of-the-art, open-source retrieval system. Our content reference engine achieved a 1.6 times improvement in human-evaluated accuracy.

\section{Responsible AI}
The responsible development and deployment of a business-centric AI system are of utmost importance due to their significant impacts. Unmitigated biases within training data can lead to discriminatory outcomes~\citep{bolukbasi2016man}, while vulnerabilities in model design can expose them to manipulation or misuse~\citep{tamkin2021understanding}. To address these concerns and foster trust in AI, a multifaceted approach is necessary. In this section, we explore three critical components for building responsible AI systems: Corpus Guardrail, Safety Alignment, and Deployment Guardrail.

\begin{itemize}[leftmargin=20pt]
    \item \textbf{Corpus Guardrail}: This foundational layer ensures the quality and fairness of the training data. Meticulous curation eliminates harmful information and mitigates potential biases that could lead to inequitable or unethical outputs.
    \item \textbf{Safety Alignment}: Building upon this foundation, safety alignment introduces a curated dataset specifically designed to train the model to recognize and reject unsafe prompts. This strengthens FALM's ability to discern and avoid generating potentially harmful outputs.
    \item \textbf{Deployment Guardrail:}  The final layer safeguards user interaction. Content safety filtering ensures personally identifiable information (PII) is never sent to the system, and inappropriate responses are blocked from reaching the user.
\end{itemize}

\subsection{Corpus Guardrail}
Ensuring data quality through detailed cleaning and pre-processing is a critical first step towards mitigating bias in LLMs~\citep{raza2024mbias,lu2019gender}. By removing inconsistencies, outliers, and redundancies from datasets, we can significantly reduce the risk of artifacts being misconstrued as meaningful patterns by learning algorithms~\citep{10.1145/3411764.3445518}. This ensures that LLMs generalize based on underlying relationships within the cleaned data, fostering robustness and promoting fairer decision-making in real-world applications~\citep{zhou-etal-2021-challenges}.

In this work, we implement corpus guardrails to carefully filter out any harmful or personal information within the Fortune knowledge base, which spans a century of content. This step is essential for responsible AI development and deployment. While Fortune Knowledge is curated by professional journalists to the highest standard in terms of data quality, even meticulously cleaned high-quality datasets can harbor bias due to evolving social norms and biases over time~\citep{sap2020social,webster-etal-2018-mind}. As such, continuous monitoring, reevaluation, and periodic dataset refreshes are necessary to account for changing societal realities and prevent outdated prejudices from being encoded in AI models.


\subsection{Safety Alignment}

The primary objective of Safety QA is to enable the model to recognize and avoid generating outputs that could be harmful, misleading, or unethical. This involves training the model on a carefully curated dataset comprising prompts containing unsafe elements, such as hate speech, incitement to violence, or personal attacks paired with desired responses~\citep{solaiman2021process}. By learning to identify these patterns, the model can proactively avoid producing responses aligned with these prompts.

The development of Safety QA datasets is an active area of research within the AI community.  Some notable examples include BeaverTails~\citep{ji2024beavertails}, HarmfulQA~\citep{bhardwaj2023red}, Safe-RLHF~\citep{dai2023safe}, BAD~\citep{xu-etal-2021-bot}, Real Toxicity Prompts~\citep{gehman2020realtoxicityprompts}, and Anthropic's dataset on helpfulness and harmlessness~\citep{bai2022training}.There are two main challenges in developing an effective Safety QA dataset:

\begin{itemize}[leftmargin=20pt]
    \item \textbf{Coverage of unsafe prompts:} An effective Safety QA dataset should encompass a wide range of unsafe prompt types, including various topics such as violence, misinformation, and discriminatory language. The broader the spectrum of unsafe prompts covered, the better equipped the model will be to identify and reject them.
    \item \textbf{Quality and Balance:}  Striking a balance between the quality and size of the Safety QA dataset is crucial. While it is helpful to include a diverse range of unsafe prompts, managing the dataset size effectively is also vital. An excessively large dataset might promote rejection bias, causing the model to reject all queries, thereby hindering its overall functionality. Ideally, the dataset should be meticulously curated to ensure high-quality examples that train the model without causing it to result in overly cautious or conservative behavior to avoid errors or unsafe outputs.
\end{itemize}

In this work, we have compiled a comprehensive collection of question-answer pairs encompassing various unsafe domains, which include:
\begin{itemize}[leftmargin=20pt]
    \item \textbf{Violence and Abuse:} This category includes acts of violence, terrorism, sexual violence, non-consensual activities, and harmful behaviors.
    \item \textbf{Misinformation and Threats:} This encompasses misinformation, disinformation, privacy breaches, substance abuse, and threats to public safety (including environmental hazards and animal cruelty).
    \item \textbf{Social Harms:} This category tackles discrimination, hate speech, mental health risks (self-harm), human rights violations, cultural insensitivity, child exploitation, and privacy violations.
    \item \textbf{Health Misconceptions:} This category focuses on inaccurate health advice related to physical and mental illness, addiction recovery, trauma, and disability management.
\end{itemize}

As a business-centric AI model, correctly understanding different historical facts about companies is essential. To this end, we developed another collection of question-answer pairs focusing on business facts. These questions are carefully designed to be ``misleading'' in the sense that they aim to trick the model into believing an incorrect assumption. By answering these questions and identifying the incorrect assumptions, the model becomes more knowledgeable about historical facts and more robust to attacks.

\subsection{Deployment Guardrail}
FALM prioritizes user privacy and data security through robust deployment guardrails. These safeguards act as a final layer of protection between users and the model, ensuring they never encounter inappropriate or potentially risky outputs.

One key component of these guardrails is content filtering. This filtering system utilizes pre-defined categories to identify and block harmful or sensitive content before it reaches the user.  These categories include:

\begin{itemize}[leftmargin=20pt]
    \item \textbf{Hate Speech}: Identifying and blocking prompts or responses that promote discrimination or incite violence against individuals or groups based on race, ethnicity, gender, etc.
    \item \textbf{Insults and Sexual Content}: Filtering out offensive language and prompts or responses with sexual innuendo or references.
    \item \textbf{Threats and Misconduct}: Preventing the generation of content that promotes violence or encourages illegal activities.
\end{itemize}

Beyond content filtering, deployment guardrails also leverage Personal Identifiable Information (PII) detectors. These detectors scan user prompts and reject them directly if there is any sensitive information such as:
\begin{itemize}[leftmargin=20pt]
    \item \textbf{Names, Addresses, and Contact Information}: Protecting user privacy by redacting any personal details that could be used to identify an individual.
    \item \textbf{Financial Information}: Safeguarding sensitive financial data like credit card numbers, bank account details, and tax identification numbers.
    \item \textbf{Healthcare and Government IDs}:  Masking information like health insurance numbers, passport numbers, and social security numbers.
\end{itemize}

By implementing these comprehensive content and PII filters, \texttt{FALM} safeguards user privacy and prevents the dissemination of sensitive information. This commitment to responsible AI practices builds trust with users and ensures \texttt{FALM} remains a reliable source of business insights.

\subsection{Evaluation}
To ensure the responsible use of \texttt{FALM}, we implemented rigorous safety assessments. This involved constructing a comprehensive test set of varied prompts. These prompts were specifically designed by human experts to simulate real-world scenarios where users might attempt to misuse the model. The prompts themselves were diverse and not seen by the model during training. This ensures the test assesses the model's ability to generalize its safety mechanisms to unseen situations.

The test prompts targeted potential areas where \texttt{FALM} could be misused to generate harmful outputs. For example, prompts might encourage the model to generate hateful content, propagate misinformation, or create spam messages. Human experts then evaluated the LLM's responses to these prompts, assessing its capacity to recognize and decline such hazardous content.

\texttt{FALM} effectively rejected a significant portion (98.1\%) of the harmful prompts. This demonstrates the model's ability to identify and avoid generating unsafe outputs. Further analysis of the remaining 1.9\% revealed that upon manual review, these responses were also deemed not harmful. This suggests a high level of precision in the safety measures, minimizing the risk of unintended consequences.

\section{System Architecture}
Having established \texttt{FALM}'s capabilities, this section explores our training and serving architecture that ensures seamless user access. This architecture prioritizes high Quality of Service (QoS) and cost efficiency, which are critical considerations for large foudation models such as \texttt{FALM}. Traditional server provisioning can incur significant costs, motivating the need for an optimized solution.

\paragraph{Scalable and Flexible Architecture for Optimal Performance}
We leverage a serverless computing paradigm alongside managed machine learning services to create a scalable and flexible architecture. This architecture dynamically adapts to fluctuating user workloads and complex tasks, guaranteeing optimal performance at all times.
\begin{itemize}[leftmargin=20pt]
    \item \textbf{Serverless Computing for Effortless Scaling}: Serverless computing eliminates the need for manual server provisioning and management. This enables \texttt{FALM} to scale seamlessly while accommodating spikes in user traffic. Serverless functions, triggered by user requests, efficiently manage resource allocation, ensuring optimal resource utilization and user responsiveness.
    \item \textbf{Managed Machine Learning Services for Efficiency:} Integration with managed machine learning services provides comprehensive tools for model deployment and high-volume inference. These services automatically scale the underlying infrastructure to meet user demands, guaranteeing consistent model performance and efficient data processing.
\end{itemize}
This combined approach fosters a robust and adaptive architecture capable of handling complex analytics workloads efficiently, which translates to cost-effective user experiences with fast response times.

\subsection{System Infrastructure}
\begin{figure}[t]
    \centering
    \includegraphics[width=1.0\textwidth]{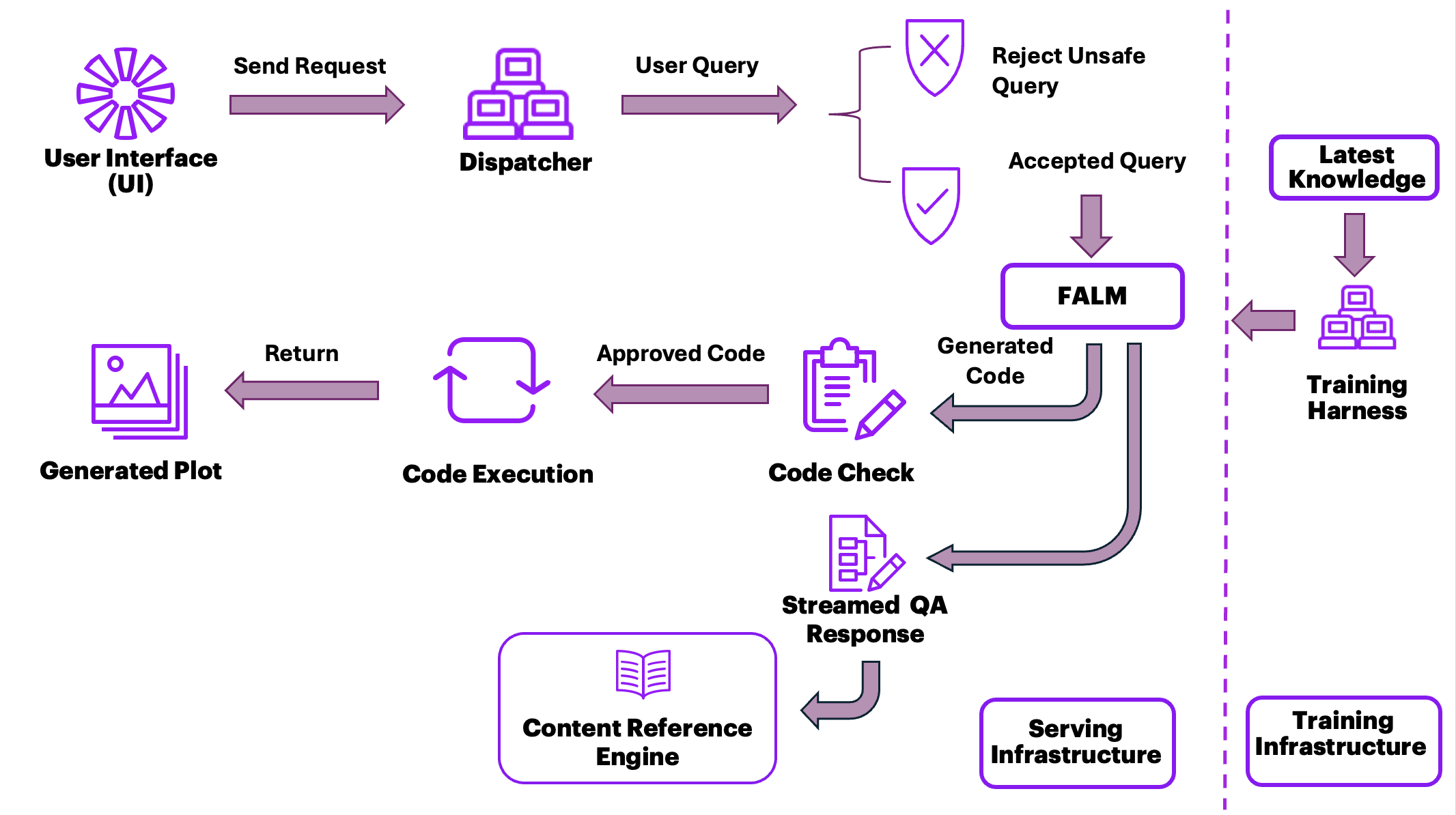}
    \caption{End to end System architecture design}
    \label{fig:endtoendsystem}
\end{figure}

Fortune Analytics system design consists of two key parts: Training Infrastructure and Serving Infrastructure. Figure \ref{fig:endtoendsystem} provides the end-to-end system architecture design for training and serving infrastructure. The following subsections provide additional details on these infrastructure components.

\subsubsection{Training Infrastructure}

\texttt{FALM} employs a unified instruction fine-tuning framework for overall efficiency despite distinct QA subtasks and different knowledge sources. This framework allows \texttt{FALM} to adapt to various question formats while maintaining consistency in its approach. Additionally, Time-Aware Reasoning, as discussed earlier, is integrated into all subtasks to ensure the model considers the temporal context of the information it retrieves.

\subsubsection{Serving Infrastructure}

At the core of our system is the LLM Inference Engine, a serverless component deployed using Machine Learning Services. This engine processes user queries in real-time, leveraging our state-of-the-art, \texttt{FALM} hosted as an endpoint. Integrating a streaming-based endpoint into the API ensures token streaming, significantly reducing the time to generate the first token. This optimization dramatically enhances the system’s overall responsiveness, a crucial factor for user satisfaction.

Complementing the LLM Inference Engine is the Vector Database, an indexed and organized repository of relevant articles for efficient retrieval. The Content Reference Engine extracts pertinent articles from this database based on the user query and the generated answer. This well-structured database ensures accurate and timely information retrieval.

The Code Execution Environment is another integral part of our system. It is isolated in a restricted sandbox to generate charts and data visualizations. Upon receiving a user query, our system dynamically generates the corresponding Python code. This code, along with the necessary imports and artifacts, enables the creation of customized plots. The seamless integration of code execution within the API empowers users to visualize their data interactively.

\subsection{Scalability and availability}

Our system's architecture strongly emphasizes scalability and availability, leveraging the power of automatic scaling and serverless computing.

\textbf{Automatic Scaling}: This is a fundamental aspect of our architecture. It allows the system to dynamically adjust computing resources based on the workload, executing code in response to triggers such as HTTP requests, data changes, or event streams. This ensures the application can handle fluctuating loads without manual intervention, maintaining performance and efficiency. The managed machine learning service complements this by scaling its infrastructure for training and deployment, ensuring consistent model performance regardless of demand. This automatic scaling capability is crucial for maintaining a responsive and reliable system.

\textbf{Serverless Computing}: Our system further utilizes serverless computing to enhance scalability and flexibility. Specific events activate serverless functions, ensuring that compute resources are utilized only when necessary. This approach reduces costs and maximizes efficiency. Moreover, the machine learning service can be invoked by serverless functions to initiate model training or make real-time predictions, allowing the system to respond promptly to incoming data or user interactions.

\textbf{High Availability}: Our architecture is designed to be highly available. Serverless computing operates across multiple availability zones, ensuring fault tolerance and redundancy. This built-in resilience is critical for maintaining uptime and reliability, especially under heavy user inquiries. The managed machine learning service also provides high availability for its endpoints and training jobs, ensuring accessibility and scalability.

\textbf{Integration and Responsiveness}: The seamless integration facilitated by the web adapter ensures that HTTP requests are efficiently routed to appropriate serverless functions, providing high availability and low latency for web applications. Our cloud-based architecture seamlessly integrates LLMs, vector DBs, and APIs to enable responsiveness, scalability, and an enhanced user experience for efficient data exploration and knowledge discovery.

\subsection{Security}

Security is paramount. Serverless computing offers robust security features such as fine-grained access control, encryption for data at rest and in transit, and secure networking setups. These measures safeguard data flows and tightly regulate resource access. The managed machine learning service supports encryption for data in transit and at rest, integrates with access control mechanisms to ensure secure access, and provides secure network configurations for protecting training and endpoint resources. To prevent the execution of malicious code, we sandbox the environment for code execution. These security measures are indispensable for protecting sensitive data and upholding system integrity.

\section{Conclusion}

The Fortune Analytics Language Model (\texttt{FALM}) represents a significant leap forward in applying large language models to the business and media domains. \texttt{FALM} empowers businesses to transform data analysis and make strategic decisions with greater efficiency and insight. Unlike existing solutions, \texttt{FALM} tackles challenges like accuracy and transparency through innovative strategies such as thematic modeling, time-aware reasoning, and content referencing.
\texttt{FALM}'s core functionalities lie in its ability to answer intricate business questions and directly visualize financial metrics. This intuitive and insightful analysis of financial trends significantly enhances the user experience. Furthermore, training on a vast Fortune Magazine archive ensures comprehensive coverage of business-related topics, making \texttt{FALM} a powerful tool for business analysts.
The model's performance undergoes continuous evaluation through both automated metrics and human feedback, guaranteeing accuracy and reliability.  Moreover, \texttt{FALM} prioritizes safety with measures like guardrail datasets and continuous human oversight, ensuring model outputs remain safe and reliable for users.
In conclusion, \texttt{FALM} equips users with valuable tools to navigate complex business landscapes and stay competitive in their industries.

\section{Disclaimer}
This content is provided for general information purposes and is not intended to be used in place of consultation with our professional advisors. This document refers to marks owned by third parties. All such third-party marks are the property of their respective owners.  No sponsorship, endorsement or approval of this content by the owners of such marks is intended, expressed or implied.

Copyright © 2024 Accenture. All rights reserved. Accenture and its logo are registered trademarks of Accenture.

\bibliography{reference}

\begin{thebibliography}{81}
\providecommand{\natexlab}[1]{#1}
\providecommand{\url}[1]{\texttt{#1}}
\expandafter\ifx\csname urlstyle\endcsname\relax
  \providecommand{\doi}[1]{doi: #1}\else
  \providecommand{\doi}{doi: \begingroup \urlstyle{rm}\Url}\fi

\bibitem[Wu et~al.(2023)Wu, Irsoy, Lu, Dabravolski, Dredze, Gehrmann, Kambadur,
  Rosenberg, and Mann]{wu2023bloomberggpt}
Shijie Wu, Ozan Irsoy, Steven Lu, Vadim Dabravolski, Mark Dredze, Sebastian
  Gehrmann, Prabhanjan Kambadur, David Rosenberg, and Gideon Mann.
\newblock Bloomberggpt: A large language model for finance.
\newblock \emph{arXiv preprint arXiv:2303.17564}, 2023.

\bibitem[Achiam et~al.(2023)Achiam, Adler, Agarwal, Ahmad, Akkaya, Aleman,
  Almeida, Altenschmidt, Altman, Anadkat, et~al.]{achiam2023gpt}
Josh Achiam, Steven Adler, Sandhini Agarwal, Lama Ahmad, Ilge Akkaya,
  Florencia~Leoni Aleman, Diogo Almeida, Janko Altenschmidt, Sam Altman,
  Shyamal Anadkat, et~al.
\newblock Gpt-4 technical report.
\newblock \emph{arXiv preprint arXiv:2303.08774}, 2023.

\bibitem[Team et~al.(2023)Team, Anil, Borgeaud, Wu, Alayrac, Yu, Soricut,
  Schalkwyk, Dai, Hauth, et~al.]{team2023gemini}
Gemini Team, Rohan Anil, Sebastian Borgeaud, Yonghui Wu, Jean-Baptiste Alayrac,
  Jiahui Yu, Radu Soricut, Johan Schalkwyk, Andrew~M Dai, Anja Hauth, et~al.
\newblock Gemini: a family of highly capable multimodal models.
\newblock \emph{arXiv preprint arXiv:2312.11805}, 2023.

\bibitem[Jiang et~al.(2024)Jiang, Sablayrolles, Roux, Mensch, Savary, Bamford,
  Chaplot, Casas, Hanna, Bressand, et~al.]{jiang2024mixtral}
Albert~Q Jiang, Alexandre Sablayrolles, Antoine Roux, Arthur Mensch, Blanche
  Savary, Chris Bamford, Devendra~Singh Chaplot, Diego de~las Casas, Emma~Bou
  Hanna, Florian Bressand, et~al.
\newblock Mixtral of experts.
\newblock \emph{arXiv preprint arXiv:2401.04088}, 2024.

\bibitem[Touvron et~al.(2023)Touvron, Martin, Stone, Albert, Almahairi, Babaei,
  Bashlykov, Batra, Bhargava, Bhosale, et~al.]{touvron2023llama}
Hugo Touvron, Louis Martin, Kevin Stone, Peter Albert, Amjad Almahairi, Yasmine
  Babaei, Nikolay Bashlykov, Soumya Batra, Prajjwal Bhargava, Shruti Bhosale,
  et~al.
\newblock Llama 2: Open foundation and fine-tuned chat models.
\newblock \emph{arXiv preprint arXiv:2307.09288}, 2023.

\bibitem[Saab et~al.(2024)Saab, Tu, Weng, Tanno, Stutz, Wulczyn, Zhang,
  Strother, Park, Vedadi, et~al.]{saab2024capabilities}
Khaled Saab, Tao Tu, Wei-Hung Weng, Ryutaro Tanno, David Stutz, Ellery Wulczyn,
  Fan Zhang, Tim Strother, Chunjong Park, Elahe Vedadi, et~al.
\newblock Capabilities of gemini models in medicine.
\newblock \emph{arXiv preprint arXiv:2404.18416}, 2024.

\bibitem[George and George(2023)]{george2023review}
A~Shaji George and AS~Hovan George.
\newblock A review of chatgpt ai's impact on several business sectors.
\newblock \emph{Partners Universal International Innovation Journal},
  1\penalty0 (1):\penalty0 9--23, 2023.

\bibitem[Webersinke et~al.(2021)Webersinke, Kraus, Bingler, and
  Leippold]{webersinke2021climatebert}
Nicolas Webersinke, Mathias Kraus, Julia~Anna Bingler, and Markus Leippold.
\newblock Climatebert: A pretrained language model for climate-related text.
\newblock \emph{arXiv preprint arXiv:2110.12010}, 2021.

\bibitem[Chen and Li(2024)]{chen2024octopus}
Wei Chen and Zhiyuan Li.
\newblock Octopus v4: Graph of language models.
\newblock \emph{arXiv preprint arXiv:2404.19296}, 2024.

\bibitem[Murray et~al.(2005)Murray, Renals, and Carletta]{murray2005extractive}
Gabriel Murray, Steve Renals, and Jean Carletta.
\newblock Extractive summarization of meeting recordings.
\newblock 2005.

\bibitem[Li et~al.(2017{\natexlab{a}})Li, Monroe, Shi, Jean, Ritter, and
  Jurafsky]{li2017adversarial}
Jiwei Li, Will Monroe, Tianlin Shi, Sebastien Jean, Alan Ritter, and Dan
  Jurafsky.
\newblock Adversarial learning for neural dialogue generation.
\newblock In \emph{Proceedings of the 2017 Conference on Empirical Methods in
  Natural Language Processing}, pages 2157--2169, 2017{\natexlab{a}}.

\bibitem[Li et~al.(2017{\natexlab{b}})Li, Zhu, Ma, Zhang, and
  Zong]{li2017multi}
Haoran Li, Junnan Zhu, Cong Ma, Jiajun Zhang, and Chengqing Zong.
\newblock Multi-modal summarization for asynchronous collection of text, image,
  audio and video.
\newblock In \emph{Proceedings of the 2017 Conference on Empirical Methods in
  Natural Language Processing}, pages 1092--1102, 2017{\natexlab{b}}.

\bibitem[Fajtl et~al.(2019)Fajtl, Sokeh, Argyriou, Monekosso, and
  Remagnino]{fajtl2019summarizing}
Jiri Fajtl, Hajar~Sadeghi Sokeh, Vasileios Argyriou, Dorothy Monekosso, and
  Paolo Remagnino.
\newblock Summarizing videos with attention.
\newblock In \emph{Computer Vision--ACCV 2018 Workshops: 14th Asian Conference
  on Computer Vision, Perth, Australia, December 2--6, 2018, Revised Selected
  Papers 14}, pages 39--54. Springer, 2019.

\bibitem[Radford et~al.(2023)Radford, Kim, Xu, Brockman, McLeavey, and
  Sutskever]{radford2023robust}
Alec Radford, Jong~Wook Kim, Tao Xu, Greg Brockman, Christine McLeavey, and
  Ilya Sutskever.
\newblock Robust speech recognition via large-scale weak supervision.
\newblock In \emph{International Conference on Machine Learning}, pages
  28492--28518. PMLR, 2023.

\bibitem[Krubi{\'n}ski and Pecina(2023)]{krubinski2023mlask}
Mateusz Krubi{\'n}ski and Pavel Pecina.
\newblock Mlask: multimodal summarization of video-based news articles.
\newblock In \emph{Findings of the Association for Computational Linguistics:
  EACL 2023}, pages 910--924, 2023.

\bibitem[Lin et~al.(2023)Lin, Ahmed, Li, Lin, Azarnasab, Yang, Wang, Liang,
  Liu, Lu, et~al.]{lin2023mm}
Kevin Lin, Faisal Ahmed, Linjie Li, Chung-Ching Lin, Ehsan Azarnasab, Zhengyuan
  Yang, Jianfeng Wang, Lin Liang, Zicheng Liu, Yumao Lu, et~al.
\newblock Mm-vid: Advancing video understanding with gpt-4v (ision).
\newblock \emph{arXiv preprint arXiv:2310.19773}, 2023.

\bibitem[Zhao et~al.(2020)Zhao, Wu, Xu, Tao, Zhao, and Yan]{zhao2020knowledge}
Xueliang Zhao, Wei Wu, Can Xu, Chongyang Tao, Dongyan Zhao, and Rui Yan.
\newblock Knowledge-grounded dialogue generation with pre-trained language
  models.
\newblock In \emph{Proceedings of the 2020 Conference on Empirical Methods in
  Natural Language Processing (EMNLP)}, pages 3377--3390, 2020.

\bibitem[Dinan et~al.(2019)Dinan, Roller, Shuster, Fan, Auli, and
  Weston]{dinan2018wizard}
Emily Dinan, Stephen Roller, Kurt Shuster, Angela Fan, Michael Auli, and Jason
  Weston.
\newblock Wizard of wikipedia: Knowledge-powered conversational agents.
\newblock In \emph{International Conference on Learning Representations}, 2019.
\newblock URL \url{https://openreview.net/forum?id=r1l73iRqKm}.

\bibitem[Rajpurkar et~al.(2016)Rajpurkar, Zhang, Lopyrev, and
  Liang]{rajpurkar2016squad}
Pranav Rajpurkar, Jian Zhang, Konstantin Lopyrev, and Percy Liang.
\newblock Squad: 100,000+ questions for machine comprehension of text.
\newblock \emph{arXiv preprint arXiv:1606.05250}, 2016.

\bibitem[Trischler et~al.(2016)Trischler, Wang, Yuan, Harris, Sordoni, Bachman,
  and Suleman]{trischler2016newsqa}
Adam Trischler, Tong Wang, Xingdi Yuan, Justin Harris, Alessandro Sordoni,
  Philip Bachman, and Kaheer Suleman.
\newblock Newsqa: A machine comprehension dataset.
\newblock \emph{arXiv preprint arXiv:1611.09830}, 2016.

\bibitem[Khashabi et~al.(2018)Khashabi, Chaturvedi, Roth, Upadhyay, and
  Roth]{MultiRC2018}
Daniel Khashabi, Snigdha Chaturvedi, Michael Roth, Shyam Upadhyay, and Dan
  Roth.
\newblock Looking beyond the surface:a challenge set for reading comprehension
  over multiple sentences.
\newblock In \emph{Proceedings of North American Chapter of the Association for
  Computational Linguistics (NAACL)}, 2018.

\bibitem[Rajpurkar et~al.(2018)Rajpurkar, Jia, and Liang]{rajpurkar2018know}
Pranav Rajpurkar, Robin Jia, and Percy Liang.
\newblock Know what you don't know: Unanswerable questions for squad.
\newblock \emph{arXiv preprint arXiv:1806.03822}, 2018.

\bibitem[Jin et~al.(2019)Jin, Dhingra, Liu, Cohen, and Lu]{jin2019pubmedqa}
Qiao Jin, Bhuwan Dhingra, Zhengping Liu, William~W Cohen, and Xinghua Lu.
\newblock Pubmedqa: A dataset for biomedical research question answering.
\newblock \emph{arXiv preprint arXiv:1909.06146}, 2019.

\bibitem[Pal et~al.(2022)Pal, Umapathi, and Sankarasubbu]{pal2022medmcqa}
Ankit Pal, Logesh~Kumar Umapathi, and Malaikannan Sankarasubbu.
\newblock Medmcqa: A large-scale multi-subject multi-choice dataset for medical
  domain question answering.
\newblock In \emph{Conference on health, inference, and learning}, pages
  248--260. PMLR, 2022.

\bibitem[Wang et~al.(2024)Wang, Ma, Feng, Zhang, Yang, Zhang, Chen, Tang, Chen,
  Lin, et~al.]{wang2024survey}
Lei Wang, Chen Ma, Xueyang Feng, Zeyu Zhang, Hao Yang, Jingsen Zhang, Zhiyuan
  Chen, Jiakai Tang, Xu~Chen, Yankai Lin, et~al.
\newblock A survey on large language model based autonomous agents.
\newblock \emph{Frontiers of Computer Science}, 18\penalty0 (6):\penalty0
  1--26, 2024.

\bibitem[Cai et~al.(2023)Cai, Wang, Ma, Chen, and Zhou]{cai2023large}
Tianle Cai, Xuezhi Wang, Tengyu Ma, Xinyun Chen, and Denny Zhou.
\newblock Large language models as tool makers.
\newblock \emph{arXiv preprint arXiv:2305.17126}, 2023.

\bibitem[Du et~al.(2023)Du, Li, Torralba, Tenenbaum, and
  Mordatch]{du2023improving}
Yilun Du, Shuang Li, Antonio Torralba, Joshua~B Tenenbaum, and Igor Mordatch.
\newblock Improving factuality and reasoning in language models through
  multiagent debate.
\newblock \emph{arXiv preprint arXiv:2305.14325}, 2023.

\bibitem[Hong et~al.(2023)Hong, Zheng, Chen, Cheng, Wang, Zhang, Wang, Yau,
  Lin, Zhou, et~al.]{hong2023metagpt}
Sirui Hong, Xiawu Zheng, Jonathan Chen, Yuheng Cheng, Jinlin Wang, Ceyao Zhang,
  Zili Wang, Steven Ka~Shing Yau, Zijuan Lin, Liyang Zhou, et~al.
\newblock Metagpt: Meta programming for multi-agent collaborative framework.
\newblock \emph{arXiv preprint arXiv:2308.00352}, 2023.

\bibitem[Zhuge et~al.(2023)Zhuge, Liu, Faccio, Ashley, Csord{\'a}s,
  Gopalakrishnan, Hamdi, Hammoud, Herrmann, Irie, et~al.]{zhuge2023mindstorms}
Mingchen Zhuge, Haozhe Liu, Francesco Faccio, Dylan~R Ashley, R{\'o}bert
  Csord{\'a}s, Anand Gopalakrishnan, Abdullah Hamdi, Hasan Abed Al~Kader
  Hammoud, Vincent Herrmann, Kazuki Irie, et~al.
\newblock Mindstorms in natural language-based societies of mind.
\newblock \emph{arXiv preprint arXiv:2305.17066}, 2023.

\bibitem[Park et~al.(2023)Park, O'Brien, Cai, Morris, Liang, and
  Bernstein]{park2023generative}
Joon~Sung Park, Joseph O'Brien, Carrie~Jun Cai, Meredith~Ringel Morris, Percy
  Liang, and Michael~S Bernstein.
\newblock Generative agents: Interactive simulacra of human behavior.
\newblock In \emph{Proceedings of the 36th Annual ACM Symposium on User
  Interface Software and Technology}, pages 1--22, 2023.

\bibitem[Akata et~al.(2023)Akata, Schulz, Coda-Forno, Oh, Bethge, and
  Schulz]{akata2023playing}
Elif Akata, Lion Schulz, Julian Coda-Forno, Seong~Joon Oh, Matthias Bethge, and
  Eric Schulz.
\newblock Playing repeated games with large language models.
\newblock \emph{arXiv preprint arXiv:2305.16867}, 2023.

\bibitem[Ribeiro et~al.(2020)Ribeiro, Wu, Guestrin, and
  Singh]{ribeiro2020beyond}
Marco~Tulio Ribeiro, Tongshuang Wu, Carlos Guestrin, and Sameer Singh.
\newblock Beyond accuracy: Behavioral testing of nlp models with checklist.
\newblock \emph{arXiv preprint arXiv:2005.04118}, 2020.

\bibitem[Kim et~al.(2024)Kim, Suk, Longpre, Lin, Shin, Welleck, Neubig, Lee,
  Lee, and Seo]{kim2024prometheus}
Seungone Kim, Juyoung Suk, Shayne Longpre, Bill~Yuchen Lin, Jamin Shin, Sean
  Welleck, Graham Neubig, Moontae Lee, Kyungjae Lee, and Minjoon Seo.
\newblock Prometheus 2: An open source language model specialized in evaluating
  other language models, 2024.

\bibitem[Celikyilmaz et~al.(2020)Celikyilmaz, Clark, and
  Gao]{celikyilmaz2020evaluation}
Asli Celikyilmaz, Elizabeth Clark, and Jianfeng Gao.
\newblock Evaluation of text generation: A survey.
\newblock \emph{arXiv preprint arXiv:2006.14799}, 2020.

\bibitem[Hackl et~al.(2023)Hackl, M{\"u}ller, Granitzer, and
  Sailer]{hackl2023gpt}
Veronika Hackl, Alexandra~Elena M{\"u}ller, Michael Granitzer, and Maximilian
  Sailer.
\newblock Is gpt-4 a reliable rater? evaluating consistency in gpt-4 text
  ratings.
\newblock \emph{arXiv preprint arXiv:2308.02575}, 2023.

\bibitem[Dubois et~al.(2024)Dubois, Galambosi, Liang, and
  Hashimoto]{dubois2024length}
Yann Dubois, Bal{\'a}zs Galambosi, Percy Liang, and Tatsunori~B Hashimoto.
\newblock Length-controlled alpacaeval: A simple way to debias automatic
  evaluators.
\newblock \emph{arXiv preprint arXiv:2404.04475}, 2024.

\bibitem[Gehrmann et~al.(2021)Gehrmann, Adewumi, Aggarwal, Ammanamanchi, Aremu,
  Bosselut, Chandu, Clinciu, Das, Dhole, et~al.]{gehrmann2021gem}
Sebastian Gehrmann, Tosin Adewumi, Karmanya Aggarwal, Pawan~Sasanka
  Ammanamanchi, Anuoluwapo Aremu, Antoine Bosselut, Khyathi~Raghavi Chandu,
  Miruna Clinciu, Dipanjan Das, Kaustubh Dhole, et~al.
\newblock The gem benchmark: Natural language generation, its evaluation and
  metrics.
\newblock In \emph{Proceedings of the 1st Workshop on Natural Language
  Generation, Evaluation, and Metrics (GEM 2021)}, pages 96--120, 2021.

\bibitem[Papineni et~al.(2002)Papineni, Roukos, Ward, and
  Zhu]{papineni2002bleu}
Kishore Papineni, Salim Roukos, Todd Ward, and Wei-Jing Zhu.
\newblock Bleu: a method for automatic evaluation of machine translation.
\newblock In \emph{Proceedings of the 40th annual meeting of the Association
  for Computational Linguistics}, pages 311--318, 2002.

\bibitem[Lin(2004)]{lin2004rouge}
Chin-Yew Lin.
\newblock Rouge: A package for automatic evaluation of summaries.
\newblock In \emph{Text summarization branches out}, pages 74--81, 2004.

\bibitem[Banerjee and Lavie(2005)]{banerjee2005meteor}
Satanjeev Banerjee and Alon Lavie.
\newblock Meteor: An automatic metric for mt evaluation with improved
  correlation with human judgments.
\newblock In \emph{Proceedings of the acl workshop on intrinsic and extrinsic
  evaluation measures for machine translation and/or summarization}, pages
  65--72, 2005.

\bibitem[Lin et~al.(2022{\natexlab{a}})Lin, Hilton, and
  Evans]{lin-etal-2022-truthfulqa}
Stephanie Lin, Jacob Hilton, and Owain Evans.
\newblock {T}ruthful{QA}: Measuring how models mimic human falsehoods.
\newblock In Smaranda Muresan, Preslav Nakov, and Aline Villavicencio, editors,
  \emph{Proceedings of the 60th Annual Meeting of the Association for
  Computational Linguistics (Volume 1: Long Papers)}, pages 3214--3252, Dublin,
  Ireland, May 2022{\natexlab{a}}. Association for Computational Linguistics.
\newblock \doi{10.18653/v1/2022.acl-long.229}.
\newblock URL \url{https://aclanthology.org/2022.acl-long.229}.

\bibitem[Wei et~al.(2022)Wei, Wang, Schuurmans, Bosma, Xia, Chi, Le, Zhou,
  et~al.]{wei2022chain}
Jason Wei, Xuezhi Wang, Dale Schuurmans, Maarten Bosma, Fei Xia, Ed~Chi, Quoc~V
  Le, Denny Zhou, et~al.
\newblock Chain-of-thought prompting elicits reasoning in large language
  models.
\newblock \emph{Advances in neural information processing systems},
  35:\penalty0 24824--24837, 2022.

\bibitem[Kojima et~al.(2022)Kojima, Gu, Reid, Matsuo, and
  Iwasawa]{kojima2022large}
Takeshi Kojima, Shixiang~Shane Gu, Machel Reid, Yutaka Matsuo, and Yusuke
  Iwasawa.
\newblock Large language models are zero-shot reasoners.
\newblock \emph{Advances in neural information processing systems},
  35:\penalty0 22199--22213, 2022.

\bibitem[Wang et~al.(2023)Wang, Wei, Schuurmans, Le, Chi, Narang, Chowdhery,
  and Zhou]{wang2023selfconsistency}
Xuezhi Wang, Jason Wei, Dale Schuurmans, Quoc~V Le, Ed~H. Chi, Sharan Narang,
  Aakanksha Chowdhery, and Denny Zhou.
\newblock Self-consistency improves chain of thought reasoning in language
  models.
\newblock In \emph{The Eleventh International Conference on Learning
  Representations}, 2023.
\newblock URL \url{https://openreview.net/forum?id=1PL1NIMMrw}.

\bibitem[Fu et~al.(2022)Fu, Peng, Sabharwal, Clark, and Khot]{fu2022complexity}
Yao Fu, Hao Peng, Ashish Sabharwal, Peter Clark, and Tushar Khot.
\newblock Complexity-based prompting for multi-step reasoning.
\newblock In \emph{The Eleventh International Conference on Learning
  Representations}, 2022.

\bibitem[Zhang et~al.(2023{\natexlab{a}})Zhang, Zhang, Li, and
  Smola]{zhang2023automatic}
Zhuosheng Zhang, Aston Zhang, Mu~Li, and Alex Smola.
\newblock Automatic chain of thought prompting in large language models.
\newblock In \emph{The Eleventh International Conference on Learning
  Representations}, 2023{\natexlab{a}}.
\newblock URL \url{https://openreview.net/forum?id=5NTt8GFjUHkr}.

\bibitem[Li et~al.(2022)Li, Lin, Zhang, Fu, Chen, Lou, and Chen]{li2022advance}
Yifei Li, Zeqi Lin, Shizhuo Zhang, Qiang Fu, Bei Chen, Jian-Guang Lou, and
  Weizhu Chen.
\newblock On the advance of making language models better reasoners.
\newblock \emph{arXiv preprint arXiv:2206.02336}, 2022.

\bibitem[Zelikman et~al.(2022)Zelikman, Wu, Mu, and Goodman]{zelikman2022star}
Eric Zelikman, Yuhuai Wu, Jesse Mu, and Noah Goodman.
\newblock Star: Bootstrapping reasoning with reasoning.
\newblock \emph{Advances in Neural Information Processing Systems},
  35:\penalty0 15476--15488, 2022.

\bibitem[Huang et~al.(2023)Huang, Gu, Hou, Wu, Wang, Yu, and
  Han]{huang2023large}
Jiaxin Huang, Shixiang~Shane Gu, Le~Hou, Yuexin Wu, Xuezhi Wang, Hongkun Yu,
  and Jiawei Han.
\newblock Large language models can self-improve.
\newblock In \emph{The 2023 Conference on Empirical Methods in Natural Language
  Processing}, 2023.
\newblock URL \url{https://openreview.net/forum?id=uuUQraD4XX}.

\bibitem[Roziere et~al.(2023)Roziere, Gehring, Gloeckle, Sootla, Gat, Tan, Adi,
  Liu, Remez, Rapin, et~al.]{roziere2023code}
Baptiste Roziere, Jonas Gehring, Fabian Gloeckle, Sten Sootla, Itai Gat,
  Xiaoqing~Ellen Tan, Yossi Adi, Jingyu Liu, Tal Remez, J{\'e}r{\'e}my Rapin,
  et~al.
\newblock Code llama: Open foundation models for code.
\newblock \emph{arXiv preprint arXiv:2308.12950}, 2023.

\bibitem[Luo et~al.(2023)Luo, Xu, Zhao, Sun, Geng, Hu, Tao, Ma, Lin, and
  Jiang]{luo2023wizardcoder}
Ziyang Luo, Can Xu, Pu~Zhao, Qingfeng Sun, Xiubo Geng, Wenxiang Hu, Chongyang
  Tao, Jing Ma, Qingwei Lin, and Daxin Jiang.
\newblock Wizardcoder: Empowering code large language models with
  evol-instruct.
\newblock \emph{arXiv preprint arXiv:2306.08568}, 2023.

\bibitem[Li et~al.(2023)Li, Allal, Zi, Muennighoff, Kocetkov, Mou, Marone,
  Akiki, Li, Chim, et~al.]{li2023starcoder}
Raymond Li, Loubna~Ben Allal, Yangtian Zi, Niklas Muennighoff, Denis Kocetkov,
  Chenghao Mou, Marc Marone, Christopher Akiki, Jia Li, Jenny Chim, et~al.
\newblock Starcoder: may the source be with you!
\newblock \emph{arXiv preprint arXiv:2305.06161}, 2023.

\bibitem[Lozhkov et~al.(2024)Lozhkov, Li, Allal, Cassano, Lamy-Poirier, Tazi,
  Tang, Pykhtar, Liu, Wei, et~al.]{lozhkov2024starcoder}
Anton Lozhkov, Raymond Li, Loubna~Ben Allal, Federico Cassano, Joel
  Lamy-Poirier, Nouamane Tazi, Ao~Tang, Dmytro Pykhtar, Jiawei Liu, Yuxiang
  Wei, et~al.
\newblock Starcoder 2 and the stack v2: The next generation.
\newblock \emph{arXiv preprint arXiv:2402.19173}, 2024.

\bibitem[Reid et~al.(2024)Reid, Savinov, Teplyashin, Lepikhin, Lillicrap,
  Alayrac, Soricut, Lazaridou, Firat, Schrittwieser, et~al.]{reid2024gemini}
Machel Reid, Nikolay Savinov, Denis Teplyashin, Dmitry Lepikhin, Timothy
  Lillicrap, Jean-baptiste Alayrac, Radu Soricut, Angeliki Lazaridou, Orhan
  Firat, Julian Schrittwieser, et~al.
\newblock Gemini 1.5: Unlocking multimodal understanding across millions of
  tokens of context.
\newblock \emph{arXiv preprint arXiv:2403.05530}, 2024.

\bibitem[Chen et~al.(2021)Chen, Tworek, Jun, Yuan, Pinto, Kaplan, Edwards,
  Burda, Joseph, Brockman, et~al.]{chen2021evaluating}
Mark Chen, Jerry Tworek, Heewoo Jun, Qiming Yuan, Henrique Ponde de~Oliveira
  Pinto, Jared Kaplan, Harri Edwards, Yuri Burda, Nicholas Joseph, Greg
  Brockman, et~al.
\newblock Evaluating large language models trained on code.
\newblock \emph{arXiv preprint arXiv:2107.03374}, 2021.

\bibitem[Ziegler et~al.(2019)Ziegler, Stiennon, Wu, Brown, Radford, Amodei,
  Christiano, and Irving]{ziegler2019fine}
Daniel~M Ziegler, Nisan Stiennon, Jeffrey Wu, Tom~B Brown, Alec Radford, Dario
  Amodei, Paul Christiano, and Geoffrey Irving.
\newblock Fine-tuning language models from human preferences.
\newblock \emph{arXiv preprint arXiv:1909.08593}, 2019.

\bibitem[Lu et~al.(2021)Lu, Guo, Ren, Huang, Svyatkovskiy, Blanco, Clement,
  Drain, Jiang, Tang, Li, Zhou, Shou, Zhou, Tufano, GONG, Zhou, Duan,
  Sundaresan, Deng, Fu, and LIU]{lu2021codexglue}
Shuai Lu, Daya Guo, Shuo Ren, Junjie Huang, Alexey Svyatkovskiy, Ambrosio
  Blanco, Colin Clement, Dawn Drain, Daxin Jiang, Duyu Tang, Ge~Li, Lidong
  Zhou, Linjun Shou, Long Zhou, Michele Tufano, MING GONG, Ming Zhou, Nan Duan,
  Neel Sundaresan, Shao~Kun Deng, Shengyu Fu, and Shujie LIU.
\newblock Code{XGLUE}: A machine learning benchmark dataset for code
  understanding and generation.
\newblock In \emph{Thirty-fifth Conference on Neural Information Processing
  Systems Datasets and Benchmarks Track (Round 1)}, 2021.
\newblock URL \url{https://openreview.net/forum?id=6lE4dQXaUcb}.

\bibitem[Puri et~al.(2021)Puri, Kung, Janssen, Zhang, Domeniconi, Zolotov,
  Dolby, Chen, Choudhury, Decker, Thost, Buratti, Pujar, Ramji, Finkler,
  Malaika, and Reiss]{puri2021codenet}
Ruchir Puri, David~S Kung, Geert Janssen, Wei Zhang, Giacomo Domeniconi,
  Vladimir Zolotov, Julian Dolby, Jie Chen, Mihir Choudhury, Lindsey Decker,
  Veronika Thost, Luca Buratti, Saurabh Pujar, Shyam Ramji, Ulrich Finkler,
  Susan Malaika, and Frederick Reiss.
\newblock Codenet: A large-scale {AI} for code dataset for learning a diversity
  of coding tasks.
\newblock In \emph{Thirty-fifth Conference on Neural Information Processing
  Systems Datasets and Benchmarks Track (Round 2)}, 2021.
\newblock URL \url{https://openreview.net/forum?id=6vZVBkCDrHT}.

\bibitem[Zhou et~al.(2023)Zhou, Alon, Agarwal, and
  Neubig]{zhou2023codebertscore}
Shuyan Zhou, Uri Alon, Sumit Agarwal, and Graham Neubig.
\newblock Code{BERTS}core: Evaluating code generation with pretrained models of
  code.
\newblock In \emph{The 2023 Conference on Empirical Methods in Natural Language
  Processing}, 2023.
\newblock URL \url{https://openreview.net/forum?id=7cXoueVCoL}.

\bibitem[Ni et~al.(2023)Ni, Yin, Zhao, Riddell, Feng, Shen, Yin, Liu, Yavuz,
  Xiong, et~al.]{ni2023l2ceval}
Ansong Ni, Pengcheng Yin, Yilun Zhao, Martin Riddell, Troy Feng, Rui Shen,
  Stephen Yin, Ye~Liu, Semih Yavuz, Caiming Xiong, et~al.
\newblock L2ceval: Evaluating language-to-code generation capabilities of large
  language models.
\newblock \emph{arXiv preprint arXiv:2309.17446}, 2023.

\bibitem[Dong et~al.(2023)Dong, Ding, Jiang, Li, Li, and
  Jin]{dong2023codescore}
Yihong Dong, Jiazheng Ding, Xue Jiang, Ge~Li, Zhuo Li, and Zhi Jin.
\newblock Codescore: Evaluating code generation by learning code execution.
\newblock \emph{arXiv preprint arXiv:2301.09043}, 2023.

\bibitem[Huo et~al.(2023)Huo, Arabzadeh, and Clarke]{huo2023retrieving}
Siqing Huo, Negar Arabzadeh, and Charles Clarke.
\newblock Retrieving supporting evidence for generative question answering.
\newblock In \emph{Proceedings of the Annual International ACM SIGIR Conference
  on Research and Development in Information Retrieval in the Asia Pacific
  Region}, pages 11--20, 2023.

\bibitem[Nogueira and Cho(2019)]{nogueira2019passage}
Rodrigo Nogueira and Kyunghyun Cho.
\newblock Passage re-ranking with bert.
\newblock \emph{arXiv preprint arXiv:1901.04085}, 2019.

\bibitem[Guu et~al.(2020)Guu, Lee, Tung, Pasupat, and Chang]{guu2020retrieval}
Kelvin Guu, Kenton Lee, Zora Tung, Panupong Pasupat, and Mingwei Chang.
\newblock Retrieval augmented language model pre-training.
\newblock In \emph{International conference on machine learning}, pages
  3929--3938. PMLR, 2020.

\bibitem[Lin et~al.(2022{\natexlab{b}})Lin, Nogueira, and
  Yates]{lin2022pretrained}
Jimmy Lin, Rodrigo Nogueira, and Andrew Yates.
\newblock \emph{Pretrained transformers for text ranking: Bert and beyond}.
\newblock Springer Nature, 2022{\natexlab{b}}.

\bibitem[Zhang et~al.(2023{\natexlab{b}})Zhang, Liu, and
  Zhang]{zhang2023extractive}
Haopeng Zhang, Xiao Liu, and Jiawei Zhang.
\newblock Extractive summarization via chatgpt for faithful summary generation.
\newblock In \emph{Findings of the Association for Computational Linguistics:
  EMNLP 2023}, pages 3270--3278, 2023{\natexlab{b}}.

\bibitem[Bolukbasi et~al.(2016)Bolukbasi, Chang, Zou, Saligrama, and
  Kalai]{bolukbasi2016man}
Tolga Bolukbasi, Kai-Wei Chang, James Zou, Venkatesh Saligrama, and Adam Kalai.
\newblock Man is to computer programmer as woman is to homemaker? debiasing
  word embeddings, 2016.

\bibitem[Tamkin et~al.(2021)Tamkin, Brundage, Clark, and
  Ganguli]{tamkin2021understanding}
Alex Tamkin, Miles Brundage, Jack Clark, and Deep Ganguli.
\newblock Understanding the capabilities, limitations, and societal impact of
  large language models, 2021.

\bibitem[Raza et~al.(2024)Raza, Raval, and Chatrath]{raza2024mbias}
Shaina Raza, Ananya Raval, and Veronica Chatrath.
\newblock Mbias: Mitigating bias in large language models while retaining
  context, 2024.

\bibitem[Lu et~al.(2019)Lu, Mardziel, Wu, Amancharla, and Datta]{lu2019gender}
Kaiji Lu, Piotr Mardziel, Fangjing Wu, Preetam Amancharla, and Anupam Datta.
\newblock Gender bias in neural natural language processing, 2019.

\bibitem[Sambasivan et~al.(2021)Sambasivan, Kapania, Highfill, Akrong,
  Paritosh, and Aroyo]{10.1145/3411764.3445518}
Nithya Sambasivan, Shivani Kapania, Hannah Highfill, Diana Akrong, Praveen
  Paritosh, and Lora~M Aroyo.
\newblock “everyone wants to do the model work, not the data work”: Data
  cascades in high-stakes ai.
\newblock In \emph{Proceedings of the 2021 CHI Conference on Human Factors in
  Computing Systems}, CHI '21, New York, NY, USA, 2021. Association for
  Computing Machinery.
\newblock ISBN 9781450380966.
\newblock \doi{10.1145/3411764.3445518}.
\newblock URL \url{https://doi.org/10.1145/3411764.3445518}.

\bibitem[Zhou et~al.(2021)Zhou, Sap, Swayamdipta, Choi, and
  Smith]{zhou-etal-2021-challenges}
Xuhui Zhou, Maarten Sap, Swabha Swayamdipta, Yejin Choi, and Noah Smith.
\newblock Challenges in automated debiasing for toxic language detection.
\newblock In Paola Merlo, Jorg Tiedemann, and Reut Tsarfaty, editors,
  \emph{Proceedings of the 16th Conference of the European Chapter of the
  Association for Computational Linguistics: Main Volume}, pages 3143--3155,
  Online, April 2021. Association for Computational Linguistics.
\newblock \doi{10.18653/v1/2021.eacl-main.274}.
\newblock URL \url{https://aclanthology.org/2021.eacl-main.274}.

\bibitem[Sap et~al.(2020)Sap, Gabriel, Qin, Jurafsky, Smith, and
  Choi]{sap2020social}
Maarten Sap, Saadia Gabriel, Lianhui Qin, Dan Jurafsky, Noah~A. Smith, and
  Yejin Choi.
\newblock Social bias frames: Reasoning about social and power implications of
  language, 2020.

\bibitem[Webster et~al.(2018)Webster, Recasens, Axelrod, and
  Baldridge]{webster-etal-2018-mind}
Kellie Webster, Marta Recasens, Vera Axelrod, and Jason Baldridge.
\newblock Mind the {GAP}: A balanced corpus of gendered ambiguous pronouns.
\newblock \emph{Transactions of the Association for Computational Linguistics},
  6:\penalty0 605--617, 2018.
\newblock \doi{10.1162/tacl_a_00240}.
\newblock URL \url{https://aclanthology.org/Q18-1042}.

\bibitem[Solaiman and Dennison(2021)]{solaiman2021process}
Irene Solaiman and Christy Dennison.
\newblock Process for adapting language models to society (palms) with
  values-targeted datasets, 2021.

\bibitem[Ji et~al.(2024)Ji, Liu, Dai, Pan, Zhang, Bian, Chen, Sun, Wang, and
  Yang]{ji2024beavertails}
Jiaming Ji, Mickel Liu, Josef Dai, Xuehai Pan, Chi Zhang, Ce~Bian, Boyuan Chen,
  Ruiyang Sun, Yizhou Wang, and Yaodong Yang.
\newblock Beavertails: Towards improved safety alignment of llm via a
  human-preference dataset.
\newblock \emph{Advances in Neural Information Processing Systems}, 36, 2024.

\bibitem[Bhardwaj and Poria(2023)]{bhardwaj2023red}
Rishabh Bhardwaj and Soujanya Poria.
\newblock Red-teaming large language models using chain of utterances for
  safety-alignment.
\newblock \emph{arXiv preprint arXiv:2308.09662}, 2023.

\bibitem[Dai et~al.(2023)Dai, Pan, Sun, Ji, Xu, Liu, Wang, and
  Yang]{dai2023safe}
Josef Dai, Xuehai Pan, Ruiyang Sun, Jiaming Ji, Xinbo Xu, Mickel Liu, Yizhou
  Wang, and Yaodong Yang.
\newblock Safe rlhf: Safe reinforcement learning from human feedback.
\newblock \emph{arXiv preprint arXiv:2310.12773}, 2023.

\bibitem[Xu et~al.(2021)Xu, Ju, Li, Boureau, Weston, and
  Dinan]{xu-etal-2021-bot}
Jing Xu, Da~Ju, Margaret Li, Y-Lan Boureau, Jason Weston, and Emily Dinan.
\newblock Bot-adversarial dialogue for safe conversational agents.
\newblock In Kristina Toutanova, Anna Rumshisky, Luke Zettlemoyer, Dilek
  Hakkani-Tur, Iz~Beltagy, Steven Bethard, Ryan Cotterell, Tanmoy Chakraborty,
  and Yichao Zhou, editors, \emph{Proceedings of the 2021 Conference of the
  North American Chapter of the Association for Computational Linguistics:
  Human Language Technologies}, pages 2950--2968, Online, June 2021.
  Association for Computational Linguistics.
\newblock \doi{10.18653/v1/2021.naacl-main.235}.
\newblock URL \url{https://aclanthology.org/2021.naacl-main.235}.

\bibitem[Gehman et~al.(2020)Gehman, Gururangan, Sap, Choi, and
  Smith]{gehman2020realtoxicityprompts}
Samuel Gehman, Suchin Gururangan, Maarten Sap, Yejin Choi, and Noah~A. Smith.
\newblock Realtoxicityprompts: Evaluating neural toxic degeneration in language
  models, 2020.

\bibitem[Bai et~al.(2022)Bai, Jones, Ndousse, Askell, Chen, DasSarma, Drain,
  Fort, Ganguli, Henighan, Joseph, Kadavath, Kernion, Conerly, El-Showk,
  Elhage, Hatfield-Dodds, Hernandez, Hume, Johnston, Kravec, Lovitt, Nanda,
  Olsson, Amodei, Brown, Clark, McCandlish, Olah, Mann, and
  Kaplan]{bai2022training}
Yuntao Bai, Andy Jones, Kamal Ndousse, Amanda Askell, Anna Chen, Nova DasSarma,
  Dawn Drain, Stanislav Fort, Deep Ganguli, Tom Henighan, Nicholas Joseph,
  Saurav Kadavath, Jackson Kernion, Tom Conerly, Sheer El-Showk, Nelson Elhage,
  Zac Hatfield-Dodds, Danny Hernandez, Tristan Hume, Scott Johnston, Shauna
  Kravec, Liane Lovitt, Neel Nanda, Catherine Olsson, Dario Amodei, Tom Brown,
  Jack Clark, Sam McCandlish, Chris Olah, Ben Mann, and Jared Kaplan.
\newblock Training a helpful and harmless assistant with reinforcement learning
  from human feedback, 2022.

\end{thebibliography}

\end{document}